\documentclass[10pt]{article}

\usepackage{graphicx}
\usepackage{amsmath,amssymb} 
\usepackage{color}
\usepackage[width=122mm,left=12mm,paperwidth=146mm,height=193mm,top=12mm,paperheight=217mm]{geometry}

\usepackage[numbers]{natbib}
\usepackage{hyperref}
\usepackage{cleveref}

\begin{document}
\pagestyle{plain}
\title{Towards Emotion-Based Synthetic Consciousness: Using LLMs to Estimate Emotion Probability Vectors}

\author{D.Sinclair\\
Imense Ltd\\
{\tt\small david@imense.com}
\and
W.T.Pye\\
Warwick University\\
{\tt\small willempye@gmail.com}
}
\maketitle

\begin{abstract}
  This paper shows how LLMs (Large Language Models) \cite{DBLP:journals/corr/VaswaniSPUJGKP17, Llama2}
  may be used to estimate a summary of the emotional state 
associated with piece of text. The summary of emotional state is a dictionary of words used to describe emotion together
with the probability of the word appearing after a prompt comprising the original text and an emotion eliciting tail.
Through emotion analysis of Amazon product reviews we demonstrate emotion descriptors can be mapped into a PCA
type space.
It was hoped that text descriptions of actions to improve a current text described state could also be elicited through a tail prompt.
Experiment seemed to indicate that this is not straightforward to make work.
This failure put our hoped for selection of action via choosing the best predicted outcome via comparing emotional responses
out of reach for the moment.

\end{abstract}


\section{Introduction}

Human behaviour is necessarily governed by emotion \citep{emotion-ai}. Sensed information about the world around us has to be reconciled
with our internal state and any action to be taken is chosen so as to lead to future state that seems preferable to our current
state \citep{love-definition},
where preferable means `my feeling is I would like to try the new state or the action possibly leading to a new state'.
If we are hungry we will often choose to eat. If we are very hungry we will take greater risk to acquire food.
If we are cold we will try to get warm etc.
Advertising aims to convince us a course of action will lead to more happiness. Sugary carbonated drinks do not objectively
lead to long term happiness but the known short term emotional response to eating sugar is desirable.
Sensed data about the world is tremendously diverse, often inaccurate and incomplete and required
responses have varying degrees of urgency.
The arbitration engine that processes these inputs needs to naturally cope with vagueness while appearing to provide certainty internally.
Emotions are the term we use to describe our experience of using this apparatus to make decisions.
The phrase computers do not have emotions is often wrongly used to assert that interactive computer software running
on a machine cannot ever exhibit or experience emotion.
Large Language Models (LLMs) \cite{DBLP:journals/corr/VaswaniSPUJGKP17, chatGpt4, Llama2} offer a ready means of linking a chunk of text with 
an estimated emotional state, bridging the gap between the world of text and the realm of human emotion.
LLMs have been used in focused sentiment analysis and are reported to perform adequately \cite{zhang2023sentiment}
but at the time of writing we are unaware of other researchers using probabilistic emotion dictionaries.

This paper explores the intersection of LLMs and emotions, demonstrating how these models can be harnessed
to estimate the emotional content of a piece of text. We present a novel approach to summarizing emotional states by
constructing a dictionary of emotion-related words and calculating the probabilities of these words appearing
following a prompt that includes both the original text and an emotion-eliciting tail. This methodology allows us
to quantitatively assess the emotional landscape of text.

To demonstrate our approach we choose a dictionary of 271 emotion describing words and estimate their probability of being
associated with a sections Amazon product reviews. Limited computational resources and time means we are only in a
position to publish a cursory study. It is likely that many emotion are correlated and an estimate of the dimension
of emotional space may be derivable via PCA analysis on a large sample of emotion vectors.

We discuss some of the limitations we encountered during experiment and some of the obstacles to producing and regulating
the behaviour of emotion based synthetic consciousness.

This paper is layed out as follows, section \ref{sec:blah} details the LLM and hardware used to run it, section \ref{sec:emdic}
details our choices of words to make up our emotion dictionary,  section \ref{sec:EV} covers estimating emotion probabilities
from  an LLM using  a tail prompt. Section \ref{sec:EV_example} shows results on Amazon reviews. A hint at the PCA structure with
in emotion vectors is given in \ref{sec:PCA}. Finally future directions are considered and conclusion given. 

\section{Interrogating the LLM with an Emotion Eliciting Tail Prompt}
\label{sec:blah}

In this work we used Facebook's open source LlaMa2 7 billion weight LLM as the core engine \cite{Llama2}.
It was necessary to use a LLM that allowed access to raw token probabilities after a prompt.
The model ran on a Mac Studio with 32 gigabytes of RAM.
With this combination of hardware and model it took 2 minutes to compute
the probabilities of the emotion descriptors in the emotion dictionary given below.

\subsection{The Emotion Dictionary}
\label{sec:emdic}

The English language is blessed with many words and an extensive literature providing examples of the
usage of these words in appropriate contexts.
For the purposes of LLMs, it is the context of a word that conveys it's meaning.
A reader will infer the meaning of an unfamiliar word through the
context they find the word used in, provided they understand the context. 
As an example, `The shotgun discombobulated the rabbit.' shows how meaning can be moderated by context.
The context created by a {\it tail prompt} will favour an associated class of words.
For the experiment detailed in this paper the following tail prompt was used to elicit
emotion descriptors, '{\em Reading this makes me feel}'. It is likely that specific
{\em emotion eliciting tail prompts} will favour specific sub classes of descriptor
but studying this is beyond the scope of this paper.

The following words were chosen to provide a broad sample of emotion descriptors.

\begin{verbatim}
acceptance, admiration, adoration, affection, afraid, agitation, 
agony, aggressive, alarm, alarmed, alienation, amazement, 
ambivalence, amusement, anger, anguish, annoyed anticipating, 
anxious, apathy, apprehension, arrogant, assertive, astonished, 
attentiveness, attraction, aversion, awe, 
baffled, bewildered, bitter, bitter sweetness, bliss, bored, 
brazen, brooding,
calm, carefree, careless, caring, charity, cheeky, cheerfulness, 
claustrophobic, coercive, comfortable, confident, confusion, 
contempt, content, courage, cowardly, cruelty, curiosity, 
cynicism,
dazed, dejection, delighted, demoralized, depressed, desire, 
despair, determined, disappointment, disbelief, discombobulated, 
discomfort, discontentment, disgruntled, disgust, disheartened, 
dislike, dismay, disoriented, dispirited, displeasure, 
distraction, distress, disturbed, dominant, doubt, dread, driven, 
dumbstruck, 
eagerness, ecstasy, elation, embarrassment, empathy, enchanted, 
enjoyment, enlightened, ennui, enthusiasm, envy, epiphany, 
euphoria, exasperated, excitement, expectancy, 
fascination, fear, flakey, focused, fondness, friendliness, 
fright, frustrated, fury, 
glee, gloomy, glumness, gratitude, greed, grief, grouchiness, 
grumpiness, guilt, 
happiness, hate, hatred, helpless, homesickness, hope, hopeless, 
horrified, hospitable, humiliation, humility, hurt, hysteria, 
idleness, impatient, indifference, indignant, infatuation, 
infuriated, insecurity, insightful, insulted, interest, 
intrigued, irritated, isolated, 
jealousy, joviality, joy, jubilation, kind, 
lazy, liking, loathing, lonely, longing, loopy, love, lust, 
mad, melancholy, miserable, miserliness, mixed up, modesty, moody, 
mortified, mystified, 
nasty, nauseated, negative, neglect, nervous, nostalgic, numb, 
obstinate, offended, optimistic, outrage, overwhelmed, 
panicked, paranoid, passion, patience, pensiveness, perplexed, 
persevering, pessimism, pity, pleased, pleasure, politeness, 
positive, possessive, powerless, pride, puzzled, 
rage, rash, rattled, regret, rejected, relaxed, relieved, 
reluctant, remorse, resentment, resignation, restlessness, 
revulsion, ruthless, 
sadness, satisfaction, scared, schadenfreude, scorn, self-caring, 
self-compassionate, self-confident, self-conscious, self-critical, 
self-loathing, self-motivated, self-pity, self-respecting, 
self-understanding, sentimentality, serenity, shame, shameless, 
shocked, smug, sorrow, spite, stressed, strong, stubborn, stuck, 
submissive, suffering, sullenness, surprise, suspense, suspicious, 
sympathy, 
tenderness, tension, terror, thankfulness, thrilled, tired, 
tolerance, torment, triumphant, troubled, trust, 
uncertainty, undermined, uneasiness, unhappy, unnerved, unsettled, 
unsure, upset, 
vengeful, vicious, vigilance, vulnerable, 
weak, woe, worried, worthy, wrath.
\end{verbatim}

This set of words is not intended to be complete or definitive in any way. Using the {\em tail prompt}
without restricting the return to emotion descriptors elicits general waffle responses that are not
straightforward to extract sentiment form.

\subsubsection{Estimating the emotion probability vector}
\label{sec:EV}

LlaMa2 \cite{Llama2} has been released in such a way as to allow developers to access estimated token weights
returned in response to a prompt. LlaMa2 has an internal vocabulary size of roughly 30,000 tokens.
This means that when LlaMa 2 estimates the probability of the next token in a sequence the probability vector
will have 30,000 elements. Some of the words in the emotion descriptor list are made up of more than one token
in which case forward conditional probabilities are used.

\begin{figure}[hbt]
  \label{fig:fig1}

  \centering
	\includegraphics[width=1.0\textwidth]{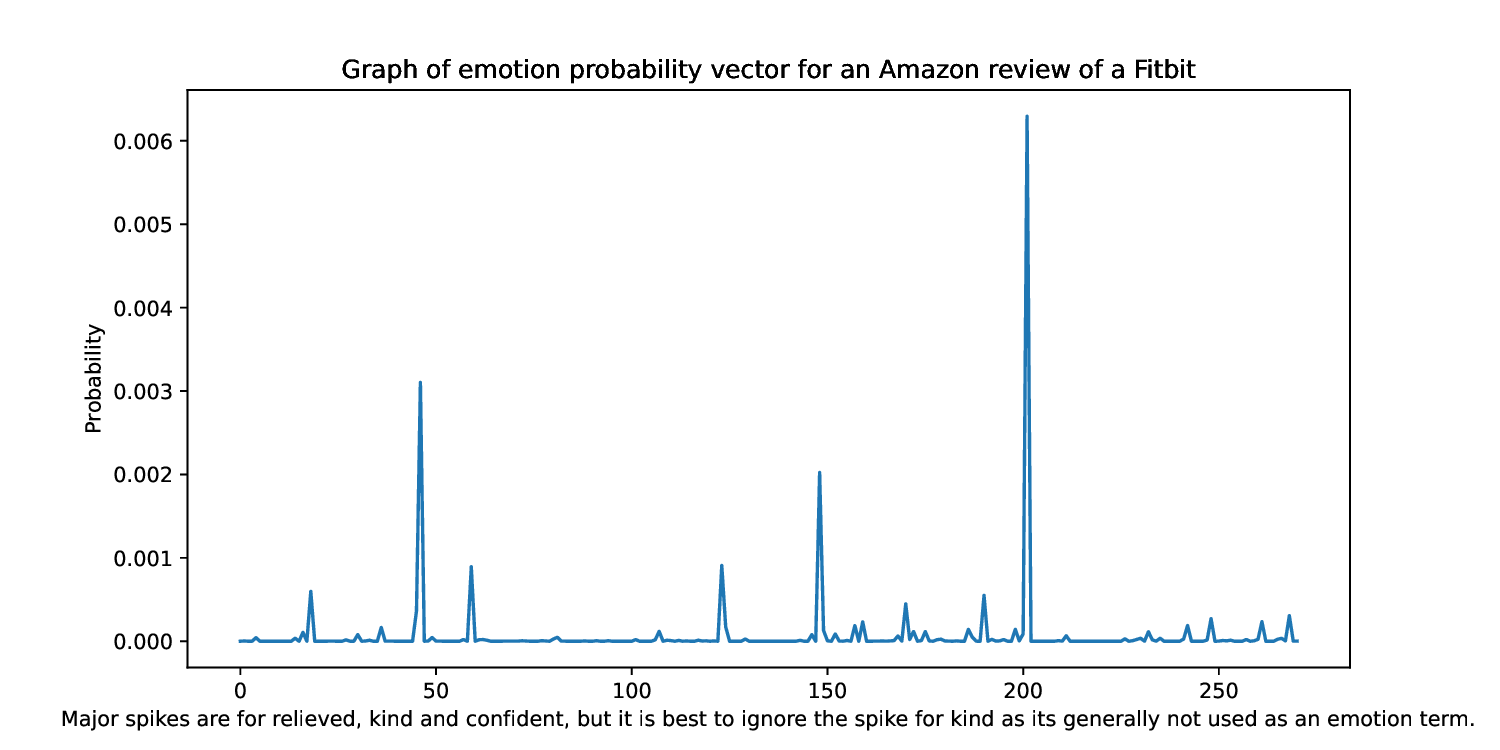}
	\caption{Example scaled emotion dictionary probabilities from an Amazon review.
          The dictionary words are ordered alphabetically.}
\end{figure}

Figure \ref{fig:fig1} shows the scaled probability distribution over words from the emotion dictionary
elicited by the tail prompt for the Amazon review text, `{\em I read a lot of negative reviews about the Fitbit inspire 2, I took a 
chance and hoped the one I ordered would be one of the great ones that 
worked. Unfortunately that was not the case. I unpacked it, charged it, 
downloaded the app. I took a walk with it on before the sun went down. I 
have the Google Fit app on my phone that tracks my steps also. The phone 
was in my jeans pocket. When I got home I compared the two, Google Fit 
said 4,458 steps, Fitbit said 1,168. Apparently Fitbit works with wrist 
motion which I don't have while pushing a walker around the neighborhood. 
I downloaded the manual and noticed you can put it on a clip (that wasn't 
included). That would work for me. So I started to scroll through the 
different features except I couldn't scroll through all of them. While 
scrolling I must have turned on the stopwatch. I couldn't turn it off. 
Then I couldn't scroll through anything except water lock feature. I had 
to turn on the water lock to get back to the stopwatch. Then the side 
buttons stopped working. I had it a total of 5 hours. I packed it up and 
started the Amazon return. I did get a full refund. Very disappointing.}'

\subsubsection{Example Emotional state from Amazon Reviews}
\label{sec:EV_example}

The text from 50 Amazon reviews of a book was borrowed from \\https://www.amazon.com/dp/B000WM9UK2.
The reviews were for the most part favourable. Example review texts include: `{\em  The Children of H\'urin is a great tragedy mixed with grace. The genealogical records at the beginning may be difficult to get through, but the story very quickly gains speed. I only gave the story four stars because of the difficulty of the first few chapters, much like Matthew’s genealogy of Christ at the beginning of his gospel. Although such records are important in both they are nonetheless difficult to get through. I still definitely would recommend this book, though, as it depicts the horrible effects of evil powers on good men and women, and yet, we must continue to resist evil no matter the tragic end. It is very telling that in Tolkien’s world, at the end of days when Morgoth returns, that it is T\'urin, a man, who ends him once and for all. Those whom Satan most destroys in this life are they who will ultimately deal his death blow, as the Revelation says, “They overcame him by the blood of the Lamb and by the word of their ‘martyria,’ [witness, testimony],” those like T\'urin, or in Scripture, those like Job.}'
\begin{figure}[hbt]
\label{fig:fig2}

\centering
    \includegraphics[width=\textwidth]{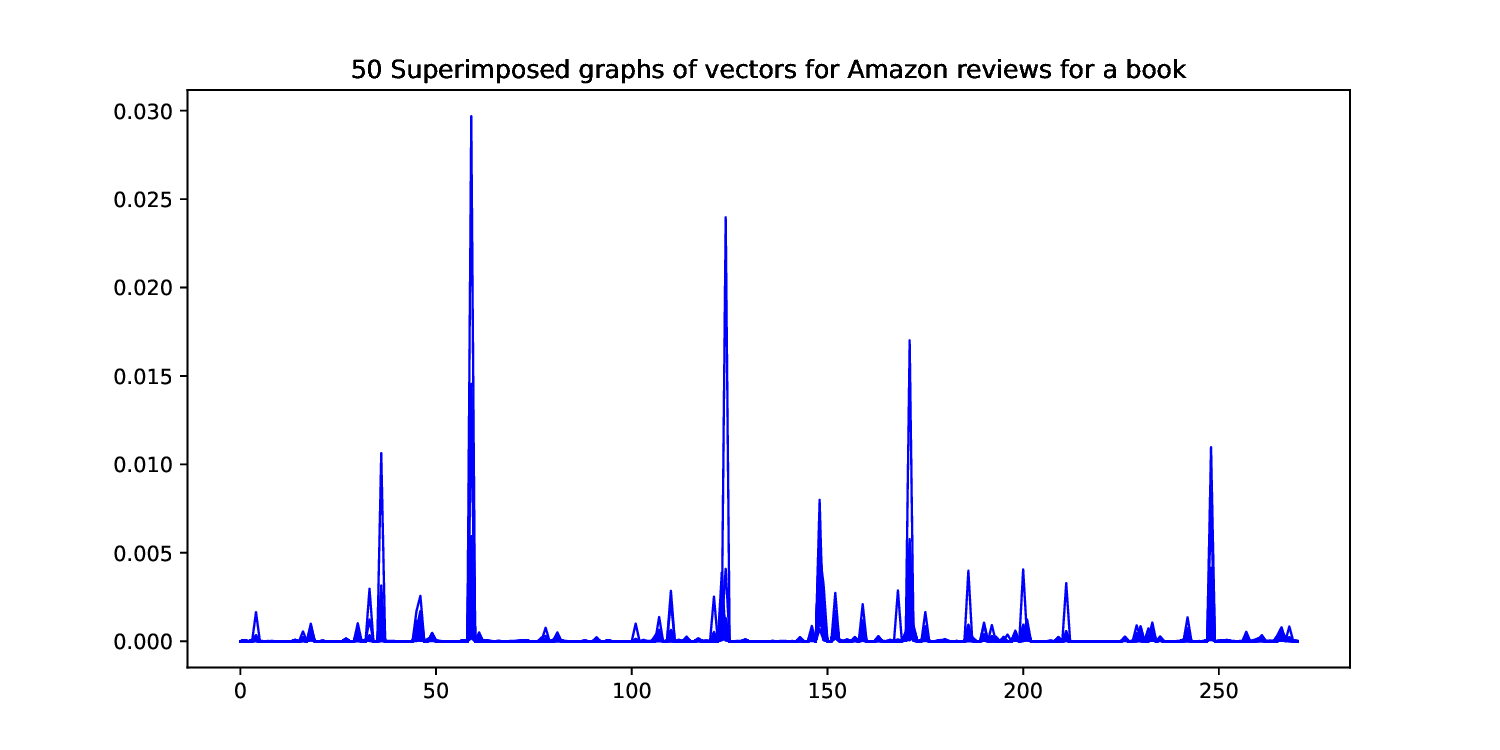}
	\caption{Superposition of emotion descriptor probabilities for 50 Amazon reviews of a book.}
\end{figure}

Figure \ref{fig:fig2} shows the emotion vectors for all 50 processed reviews.
The 10 most probable emotions experienced from purchasing the product were:
depressed,
kind,
nostalgic,
tired,
hopeless,
lonely,
hope,
calm,
lazy,
confident.

\section{PCA analysis of the Emotion of Amazon reviews}
\label{sec:PCA}

\begin{figure}[hbt]
  \label{fig:fig3}

  \centering
    \includegraphics[width=0.8\textwidth]{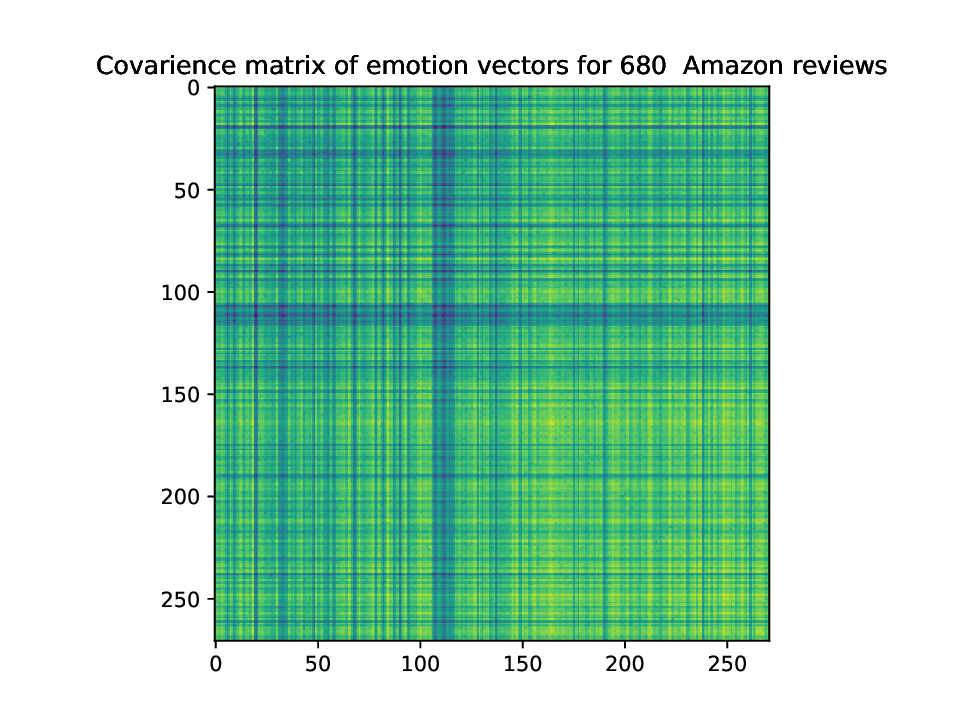}
	\caption{Co-occurance matrix derived form emotion vectors of several Amazon product reviews.}
\end{figure}

A range of Amazon products were selected and processed to estimate associated emotion vectors. Time and processing constraints mean only
680 reviews were processed. The co-occurance matrix for this data is shown in figure \ref{fig:fig3}.

\begin{figure}[hbt]
  \label{fig:fig4}

  \centering
    \includegraphics[width=0.8\textwidth]{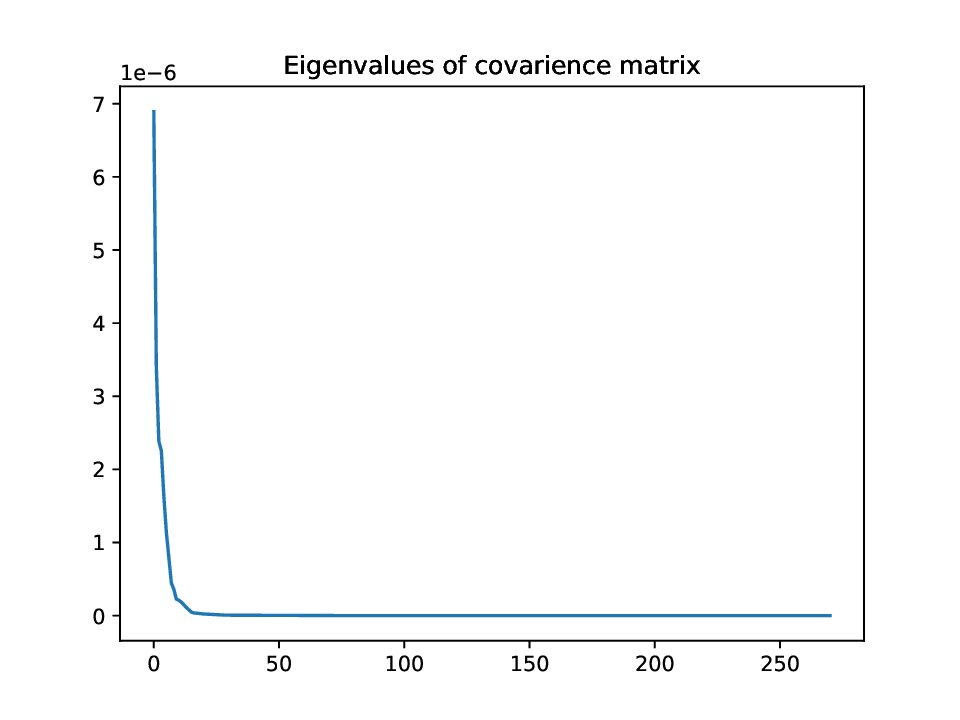}
	\caption{Sorted Eigen values of the co-occurance matrix in figure \ref{fig:fig3}.}
\end{figure}

The Eigen system of this co-occurance matrix was computed and the sorted Eigen vectors are displayed in figure \ref{fig:fig4}.
As can be see the space of emotions can be spanned by fewer than 271 emotion vectors.

\section{Future Work}
\label{sec:future}

The authors had hoped to build a skeletal self aware emotion derived synthetic consciousness.
The state of the (synthetic conscious) system is described in text.
The synthetic consciousness's perception of it's own state
is the vector of probabilities of emotion descriptors derived from one or more tail prompts
used to estimate the relevant token probabilities via the LLM associated with the system.

It was hoped that the fine grained probability vector would be useable to determine whether one text
description of current or future state was preferable to another state. This would provide a general
means of arbitrating between potentially unrelated behaviours with unrelated goals.

It was further hoped that a tail prompt could be used to elicit a text description of a
putative course of action from a LLM.
A brief series of experiments with various LLMs indicated that this was not going to work.
Example text and tail prompts included things like, `My girlfriend hates me. How can I make this better?'.
The replies read like excerpts from self help books or newspaper psychologist waffle and were not
specific enough to have a chance at creating a predicted future in text through re-insertion into the LLM.
Similar phrases appended to bad restaurant reviews elicited similarly nondescript advice.

The take home message was that the remedy proposed was too vague for the LLM to make any meaningful
prediction about the state after the advice had been taken.

This does not mean that more thoughtful prompt design will not elicit useful action prediction
hoped to improve the self perceived state of a synthetic consciousness.

\subsection{Longer Term Behaviour Regulation}
\label{sec:longer}

If synthetic consciousnesses are to have a roll in the future of humanity it would seem desirable to
endow them with a degree of empathy for living beings and a longer term view than simple
optimisation to fulfil a limited short term goal.
For example if a synthetic consciousness were to have a goal of; `{\em make money for shareholders in a company}'
it would be great if it would choose {\em not} to open an open cast coal mine and build coal fired power stations,
or `{\em take out life insurance policies on random individuals and murder them with self driving cars'}.

It has been argued that longer term altruistic behaviour in humans is moderated by love \cite{love-definition} and
a computationally feasible definition of {\em Love} is given: `{\em Love is that which prefers life}'.
Love in humans is intimately related to the production and fostering of new lives. Love seems to act to
prefer a future in which there is more life. Acting contrary to love and creating a future where there is
a wasteland with nothing living in it is generally viewed as wrong.

The advent of LLMs offers a means of creating text descriptors of predicted futures with a range of time constants.
The emotion vectors associated with predicted futures can be used to arbitrate between sort term behaviours.
Text descriptors can play a roll in behaviour regulation and a machine may act in a way that
at least in part mirrors {\em Love}. For example if an agricultural robot was invited to dump unused pesticide into a
river it might reasonably infer that this action was in principle wrong.

\section{Conclusions}

LLMs are by their nature designed to provide text strings as a response to a test prompt.
This is not always the most useful format for information to be returned in.
Internally within the LLM there exist probability distributions over tokens.
The paper presents an example of how to build part of an emotion based synthetic consciousness by
deriving the vector of emotion descriptor probabilities over a dictionary of emotional terms.
There are a range of things that can be done with this emotion probability vector including
fine grained review analysis, predicting a response to marketing messages, offence detection etc. 
It is possible that the emotion probability vector might be a step on the road to synthetic consciousness
and that it might provide a means of making robots more empathetic through allowing them to
make a prediction as to how something they might say will make the recipient feel.

If reasonable responses are desired from an LLM it might be a good policy not to train the LLM on the mad
shouting that pervades anti-social media and analogously it might be a good idea not to train young minds
similarly.

\section{Acknowledgements}
The authors acknowledge the extraordinary generosity of Meta in releasing model weights in a reasonable way for their
LlaMa2 series of pre-trained Large Language Models.

\bibliographystyle{plainnat}
{\small
\bibliography{references.bib} }

\begin{thebibliography}{6}
\providecommand{\natexlab}[1]{#1}
\providecommand{\url}[1]{\texttt{#1}}
\expandafter\ifx\csname urlstyle\endcsname\relax
  \providecommand{\doi}[1]{doi: #1}\else
  \providecommand{\doi}{doi: \begingroup \urlstyle{rm}\Url}\fi

\bibitem[AI(2023)]{chatGpt4}
Open AI.
\newblock Chatgpt-4 technical report.
\newblock 2023.
\newblock URL \url{https://arxiv.org/pdf/2303.08774.pdf}.

\bibitem[GenAI et~al.(2023)GenAI, Scialom, and Touvron]{Llama2}
Meta GenAI, Thomas Scialom, and Hugo Touvron.
\newblock Llama 2: Open foundation and fine-tuned chat models.
\newblock 2023.
\newblock URL \url{https://arxiv.org/pdf/2307.09288.pdf}.

\bibitem[Picard(1997)]{emotion-ai}
Rosalind~W. Picard.
\newblock Affective computing.
\newblock \emph{MIT Press}, 1997.

\bibitem[Strabismus(2013)]{love-definition}
J~Strabismus.
\newblock The jedi religion: Is love the force?
\newblock \emph{Amazon Kindle}, 2013.

\bibitem[Vaswani et~al.(2017)Vaswani, Shazeer, Parmar, Uszkoreit, Jones, Gomez,
  Kaiser, and Polosukhin]{DBLP:journals/corr/VaswaniSPUJGKP17}
Ashish Vaswani, Noam Shazeer, Niki Parmar, Jakob Uszkoreit, Llion Jones,
  Aidan~N. Gomez, Lukasz Kaiser, and Illia Polosukhin.
\newblock Attention is all you need.
\newblock \emph{CoRR}, abs/1706.03762, 2017.
\newblock URL \url{http://arxiv.org/abs/1706.03762}.

\bibitem[Zhang et~al.(2023)Zhang, Deng, Liu, Pan, and Bing]{zhang2023sentiment}
Wenxuan Zhang, Yue Deng, Bing Liu, Sinno~Jialin Pan, and Lidong Bing.
\newblock Sentiment analysis in the era of large language models: A reality
  check, 2023.

\end{thebibliography}






\end{document}